\documentclass{article}

\usepackage[preprint]{nips2018}

\usepackage{graphicx} 
\usepackage{subfig}
\usepackage{xspace}
\usepackage[font={small}]{caption}
\usepackage{color}
\usepackage{longtable}
\usepackage{hhline}

\usepackage{natbib}

\usepackage{algorithm}
\usepackage{algorithmic}

\usepackage{hyperref}

\usepackage{verbatim}		

\usepackage{url}		
\usepackage{fancyhdr}		
\usepackage{datetime}		
\usepackage{lastpage}		

\usepackage{amsmath}
\usepackage{amssymb}
\usepackage{amsthm}
\usepackage{xspace}

\makeatletter
\newtheorem*{rep@theorem}{\rep@title}
\newcommand{\newreptheorem}[2]{%
\newenvironment{rep#1}[1]{%
 \def\rep@title{#2 \ref{##1}}%
 \begin{rep@theorem}}%
 {\end{rep@theorem}}}
\makeatother

\newreptheorem{theorem}{Theorem}
\newreptheorem{definition}{Definition}

\newtheorem{theorem}{Theorem}[section]

\newtheorem{proposition}[theorem]{Proposition}

\newcommand{\ann}{ANN\xspace}
\newcommand{\annnp}{ANN} 
\newcommand{\anns}{ANNs\xspace}

\newcommand{\adaloss}{AdaLoss\xspace}

\newcommand{\const}{CONST\xspace}
\newcommand{\linear}{LINEAR\xspace}

\usepackage{enumitem} 
\AtBeginDocument{%
 \abovedisplayskip=4pt minus 1pt
 \abovedisplayshortskip=2pt plus 1pt
 \belowdisplayskip=4pt minus 1pt
 \belowdisplayshortskip=2pt plus 1pt
}
\setitemize{itemsep=0mm, leftmargin=5mm, topsep=0mm}
\setenumerate{itemsep=0mm, leftmargin=5mm, topsep=0mm}
\usepackage[noindentafter]{titlesec}
\titlespacing\section{0pt}{4pt plus 0pt minus 2pt}{2pt plus 0pt minus 1pt}
\titlespacing\subsection{0pt}{4pt plus 0pt minus 1pt}{1pt plus 1pt minus 1pt}
\titlespacing\subsubsection{0pt}{4pt plus 0pt minus 1pt}{1pt plus 1pt minus 1pt}

\title{Learning Anytime Predictions in Neural Networks\\ 
via Adaptive Loss Balancing  }

\author{
  Hanzhang Hu \\
  School of Computer Science\\
  Carnegie Mellon University\\
  Pittsburgh, PA 15213 \\
  \texttt{hanzhang@cs.cmu.edu} \\
  \And
  Debadeepta Dey\\
  Microsoft Research \\
  Redmond, WA 98052 \\
  \texttt{dedey@microsoft.com} \\
  \And
  Martial Hebert \\
  School of Computer Science \\
  Carnegie Mellon University \\
  Pittsburgh, PA 15213 \\
  \texttt{hebert@cs.cmu.edu}
  \And
  J. Andrew Bagnell \\
  School of Computer Science \\
  Carnegie Mellon University \\
  Pittsburgh, PA 15213 \\
  \texttt{dbagnell@cs.cmu.edu} \\
}

\begin{document}
\maketitle

\begin{abstract}

This work considers the trade-off between accuracy and test-time computational cost of deep neural networks (DNNs) via \emph{anytime} predictions from auxiliary predictions. Specifically, we optimize auxiliary losses jointly in an \emph{adaptive} weighted sum, where the weights are inversely proportional to average of each loss. 
Intuitively, this balances the losses to have the same scale.
We demonstrate theoretical considerations that motivate this approach from multiple viewpoints, including connecting it to optimizing the geometric mean of the expectation of each loss, an objective that ignores the scale of losses. 
Experimentally, the adaptive weights induce more competitive anytime predictions on multiple recognition data-sets and models than non-adaptive approaches including weighing all losses equally. In particular, anytime neural networks (\anns) can achieve the same accuracy faster using adaptive weights on a small network than using static constant weights on a large one.
For problems with high performance saturation, we also show a sequence of exponentially deepening \anns can achieve near-optimal anytime results at any budget, at the cost of a const fraction of extra computation. 
\end{abstract}


\section{Introduction}
\label{sec:introduction}

Recent years have seen advancement in visual recognition tasks
by increasingly accurate convolutional neural networks, from AlexNet~\citep{alexnet} and VGG~\citep{vgg}, to ResNet~\citep{resnet}, ResNeXt~\citep{resnext}, and DenseNet~\citep{densenet}. 
As models become more accurate and computationally expensive, it becomes more difficult for applications to choose between slow predictors with high accuracy and fast predictors with low accuracy. 
Some applications also desire multiple trade-offs between computation and accuracy, because they have computational budgets that may vary at test time. E.g., web servers for facial recognition or spam filtering may have higher load during the afternoon than at midnight.  Autonomous vehicles need faster object detection when moving rapidly than when it is stationary.  Furthermore, real-time and latency sensitive applications may desire fast predictions on easy samples and slow but accurate predictions on difficult ones. 

An \textbf{anytime predictor}~\citep{horvitz:1987,boddydean,anytime, speedboost,msdense} can automatically trade off between computation and accuracy. For each test sample, an anytime predictor produces a fast and crude initial prediction and continues to refine it as budget allows, so that at any test-time budget, the anytime predictor has a valid result for the sample, and the more budget is spent, the better the prediction. 
Anytime predictors are different from cascaded predictors~\citep{cascade,xu:14,cai:15,adaptivenn,cascade_nn} for \textbf{budgeted prediction}, which aim to minimize \textbf{average test-time computational cost} without sacrificing average accuracy: a different task (with relation to anytime prediction). Cascades achieve this by early exiting on easy samples to save computation for difficult ones, but cascades cannot incrementally improve individual samples after an exit. Furthermore, early exit policy of cascades can be combined with existing anytime predictors~\citep{adaptivenn,cascade_nn}. Hence, we consider cascades to be orthogonal to anytime predictions.

\begin{figure}
    \centering
    \subfloat[ ]{
        \includegraphics[width=0.37\linewidth, keepaspectratio]{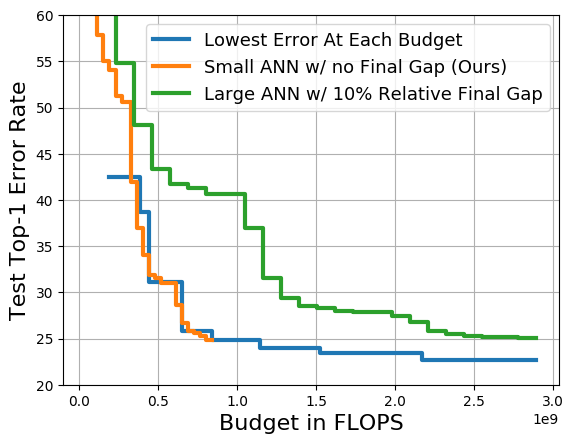}
        \label{fig:sieve_small_vs_const_large}
    }
    ~
    \subfloat[]{
        \includegraphics[width=0.37\linewidth, keepaspectratio]{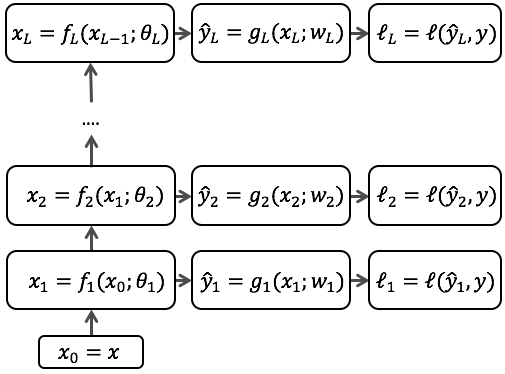}
        \label{fig:ann}
    }
    \caption{\textbf{(a)} The common \ann training strategy increases final errors from the optimal (green vs. blue), which decreases exponentially slowly. By learning to focus more on the final auxiliary losses, the proposed adaptive loss weights make a small \ann (orange) to outperform a large one (green) that has non-adaptive weights.  
    \textbf{(b)} Anytime neural networks contain auxiliary predictions and losses, $\hat{y}_i$ and $\ell_i$, for intermediate feature unit $f_i$.  }
    \vspace{-20pt}
\end{figure}

This work studies how to convert well-known DNN architectures to produce competitive anytime predictions.  
We form anytime neural networks (\anns) by appending auxiliary predictions and losses to DNNs, as we will detail in Sec.~\ref{sec:ann} and Fig.~\ref{fig:ann}. Inference-time prediction then can be stopped at the latest prediction layer that is within the budget. Note that this work deals with the case where it is \textbf{not known apriori} where the interrupt during inference time will occur. 
We define the optimal at each auxiliary loss as the result from training the \ann only for that loss to convergence. Then our objective is to have near-optimal final predictions and competitive early ones.  Near-optimal final accuracy is imperative for anytime predictors, because, as demonstrated in Fig.~\ref{fig:sieve_small_vs_const_large}, accuracy gains are often exponentially more expensive as model sizes grow, so that reducing 1\% error rate could take 50\% extra computation. Unfortunately, existing anytime predictors often optimize the anytime losses in static weighted sums~\citep{supervisednet, feedbacknet, msdense} that poorly optimize final predictions, as we will show in Sec.~\ref{sec:multi_objective} and Sec.~\ref{sec:experiment_questions}.

Instead, we optimize the losses in an \textbf{adaptive} weighted sum, where the weight of a loss is inversely proportional to the empirical mean of the loss on the training set. Intuitively, this normalizes losses to have the same scale, so that the optimization leads each loss to be about the same relative to its optimal. We provide multiple theoretical considerations to motivate such weights.
First of all, when the losses are mean square errors, our approach is maximizing the likelihood of a model where the prediction targets have Gaussian noises. Secondly, inspired by the maximum likelihood estimation, we optimize the model parameters and the loss weights jointly, with log-barriers on the weights to avoid the trivial solution of zero weights. Finally, we find the joint optimization equivalent to optimizing the geometric mean of the expected training losses, an objective that treats the relative improvement of each loss equally. Empirically, we show on multiple models and visual recognition data-sets that the proposed adaptive weights outperform natural, non-adaptive weighting schemes as follows.
We compare small \anns using our adaptive weights against \anns that are $50\sim 100\%$ larger but use non-adaptive weights. The small \anns can reach the same final accuracy as the larger ones, and reach each accuracy level faster.

Early and late accuracy in an \ann are often anti-correlated (e.g., Fig.~7 in~\citep{msdense} shows \anns with better final predictions have worse early ones). To mitigate this \emph{fundamental} issue we propose to assemble \anns of exponentially increasing depths. If \anns are near-optimal in a late fraction of their layers, the exponential ensemble only pays a constant fraction of additional computation to be near-optimal at every test-time budget. In addition, exponential ensembles outperform linear ensembles of networks, which are commonly used baselines for existing works~\citep{feedbacknet, msdense}. 
In summary our contributions are:
\begin{itemize}
    \item We derive an adaptive weight scheme for training losses in \anns from multiple theoretical considerations, and show that experimentally this scheme achieves near-optimal final accuracy \emph{and} competitive anytime ones on multiple data-sets and models.
    \item We assemble \anns of exponentially increasing depths to achieve near-optimal anytime predictions at every budget at the cost of a constant fraction of additional consumed budget. 
\end{itemize}


\vspace{-3pt}
\section{Related Works}
\vspace{-3pt}
\label{sec:background}

\textbf{Meta-algorithms for anytime and budgeted prediction.}
Anytime and budgeted prediction has a rich history in learning literature.
\citep{weinberger09feature, xu:12, xu:13b} sequentially generate features to empower the final predictor.
\citep{reyzin:11, speedboost, hu:16} apply boosting and greedy methods to order feature and predictor computation.  
\citep{timeliness, rl_anytime} form Markov Decision Processes for computation of weak predictors and features, and learn policies to order them. However, these meta-algorithms are not easily compatible with complex and accurate predictors like DNNs, because the anytime predictions without DNNs are inaccurate, and there are no intermediate results during the computation of the DNNs.
Cascade designs for budgeted prediction \citep{cascade, lefakis:10, chen:12, xu:14, cai:15, adaptive_select, adaptivenn, cascade_nn} reduce the average test-time computation by early exiting on easy samples and saving computation for difficult ones. As cascades build upon existing anytime predictors, or combine multiple predictors, they are orthogonal to learning \anns end-to-end.

\textbf{Neural networks with early auxiliary predictions.} 
Multiple works have addressed training DNNs with early auxiliary predictions for various purposes. \citep{supervisednet, inception_v4, pspnet, fractalnet} use them to regularize the networks for faster and better convergence. \citep{curriculum, feedbacknet} set the auxiliary predictions from easy to hard for curriculum learning. \citep{hed,reverse_scene_seg} make pixel level predictions in images, and find learning early predictions in coarse scales also improve the fine resolution predictions. \citep{msdense} shows the crucial importance of maintaining multi-scale features for high quality early classifications. The above works use manually-tuned static weights to combine the auxiliary losses, or change the weights only once~\citep{reverse_scene_seg}. This work proposes adaptive weights to balance the losses to the same scales online, and provides multiple theoretical motivations. We empirically show adaptive losses induce better \anns on multiple models, including the state-of-the-art anytime predictor for image recognition, MSDNet~\citep{msdense}. 

\textbf{Model compression.} 
Many  works have studied how to compress neural networks.  \citep{prune_nn, slim_nn} prune network weights and connections. \citep{binary_nn, binary_nn_eccv, squeezenet} quantize weights within networks to reduce computation and memory footprint. 
\citep{wang2017skipnet, veit2017convolutional} dynamically skip network computation based on samples.
\citep{deepreally, distillation} transfer knowledge of deep networks into shallow ones by changing the training target of shallow networks. 
These works are orthogonal to ours, because they train a separate model for each trade-off between computation and accuracy, but we train a single model to handle all possible trade-offs.


\section{Optimizing Anytime Neural Network Performance}

\label{sec:ann}

\label{sec:multi_objective}

As illustrated in Fig.~\ref{fig:ann}, a feed-forward network consists of a sequence of transformations $f_1,...,f_L$ of feature maps. Starting with the input feature map $x_0$, each subsequent feature map is generated by $x_i = f_i(x_{i-1})$.  Typical DNNs use the final feature map $x_L$ to produce predictions, and hence require the completion of the whole network for results. Anytime neural networks (\anns) instead introduce auxiliary predictions and losses using the intermediate feature maps $x_1,...,x_{L-1}$, and thus, have early predictions that are improving with computation.


\textbf{Weighted sum objective.} Let the intermediate predictions be $\hat{y}_i = g_i(x_i)$ for some function $g_i$, and let the corresponding expected loss be $\ell_i = E_{(x_0,y)\sim \mathcal{D}} [\ell(y, \hat{y}_i)]$, where $\mathcal{D}$ is the distribution of the data, and $\ell$ is some loss such as cross-entropy.  Let $\theta$ be the parameter of the \ann, and define the optimal loss at prediction $\hat{y}_i$ to be $\ell_i* = \min _{\theta}  \ell_i(\theta)$. Then the goal of anytime prediction is to seek a universal 
$
    \theta^* \in \cap _{i=1}^L \{ \theta' : \theta' = \arg \min _{\theta} \ell _i(\theta) \}.
    \label{eq:multi-objective}
$
Such an ideal $\theta^*$ does not exist in general as this is a multi-objective optimization, which only has Pareto front, a set containing all solutions such that improving one $\ell_i$ necessitates degrading others. Finding all solutions in the Pareto front for \anns is not practical or useful, since this requires training multiple models, but each \ann only runs one. Hence, following previous works on anytime models~\citep{supervisednet, feedbacknet, msdense}, we optimize the losses in a weighted sum
$
    \min _{\theta} \sum _{i=1}^L B_i \ell_i (\theta),
    \label{eq:sum-objective}
$
where $B_i$ is the weight of the loss $\ell_i$. We call the choices of $B_i$ \textit{weight schemes}.


\textbf{Static weight schemes.} Previous works often use static weight schemes as part of their formulation. \cite{supervisednet, hed, msdense} use \const scheme that sets $B_i =1$ for all $i$. \cite{feedbacknet} use \linear scheme that sets $B_1$ to $B_L$ to linearly increase from $0.25$ to $1$. However, as we will show in Sec.~\ref{sec:compare_opt}, these static schemes not only cannot adjust weights in a data and model-dependent manner, but also may significantly degrade predictions at later layers.

\begin{figure}
    \centering
    
    \subfloat[Relative Percentage Increase in Training Loss vs. depths (lower is better)]{
        \includegraphics[width=0.4\linewidth, keepaspectratio]{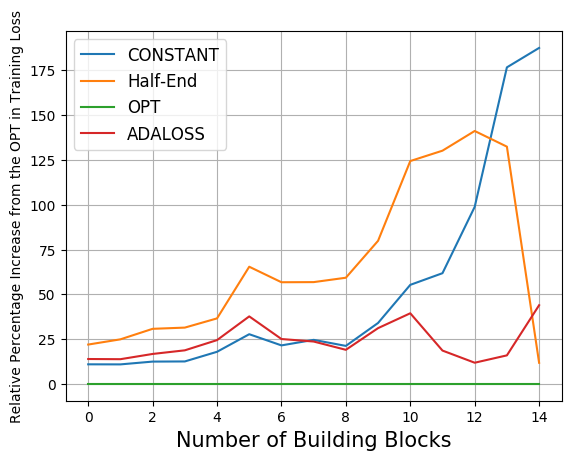}
    \label{fig:loss_cifar100}
    }
    ~
    \subfloat[Ensemble of exponentially deepening anytime neural network (EANN)]{
        \includegraphics[width=0.35\linewidth, keepaspectratio]{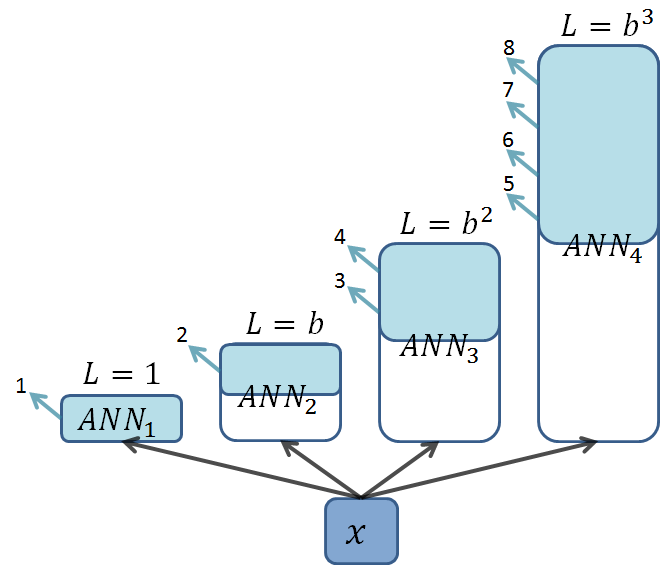}
    \label{fig:eann}
    }
    \caption{\textbf{(a)} \const scheme is increasingly worse than the optimal at deep layers. \adaloss performs about equally well on all layers in comparison to the OPT.
    \textbf{(b)} EANN computes its \anns in order of their depths. An anytime result is used if it is better than all previous ones on a validation set (layers in light blue).} 
    
    \vspace{-20pt}
\end{figure}
    

\textbf{Qualitative weight scheme comparison.} Before we formally introduce our proposed adaptive weights, we first shed light on how existing static weights suffer. We experiment with a ResNet of 15 basic residual blocks on CIFAR100~\citep{cifar} data-set (See Sec.~\ref{sec:experiment_questions} for data-set details). An anytime predictor is attached to each residual block, and we estimate the optimal performance (OPT) in training cross entropy of predictor $i$ by training a network that has weight only on $\ell_i$ to convergence. Then for each weight scheme we train an \ann to measure the relative increase in training loss at each depth $i$ from the OPT. In Fig.~\ref{fig:loss_cifar100}, we observe that the intuitive \const scheme has high relative losses in late layers. This indicates that there is not enough weights in the late layers, though losses have the same $B_i$.  We also note that balancing the weights is non-trivial. For instance, if we put half of the total weights in the final layer and distribute the other half evenly, we get the ``Half-End" scheme. As expected, the final loss is improved, but this is at the cost of significant increases of early training losses. In contrast, the adaptive weight scheme that we propose next (\adaloss), achieves roughly even relative increases in training losses automatically, and is much better than the \const scheme in the late layers.


\textbf{Adaptive Loss Balancing (\adaloss).} 
Given all losses are of the same form (cross-entropy), it may be surprising
that better performance is achieved with differing weights.
Because early features typically have less predictive power than later ones, early losses are naturally on a larger scale and possess larger gradients. Hence, if we weigh losses equally, early losses and gradients often dominate later ones, and the optimization becomes focused on the early losses.  
To automatically balance the weights among the losses of different scales, we propose an adaptive loss balancing scheme (\adaloss). Specifically, we keep an exponential average of each loss $\hat{\ell}_i$ during training, and set $B_i \propto  \frac{1}{\hat{\ell}_i}$. This is inspired by \citep{reverse_scene_seg}, which scales the losses to the same scale \emph{only once} during training, and provides a brief intuitive argument: the adaptive weights set the losses to be on the same scale. 
We next present multiple theoretical justifications for \adaloss.
 
Before considering general cases, we first consider a simple example, where the loss function $\ell(y, \hat{y}) = \Vert y - \hat{y}\Vert^2_2$ is the square loss. For this example, we model each $y|x$ to be sampled from the multiplication of $L$ independent Gaussian distributions, $\mathcal{N}(\hat{y}_i, \sigma_i^2 I)$ for $i=1,...,L$, where $\hat{y}_i(x; \theta)$ is the $i^{th}$ prediction, and $\sigma_i^2 \in \mathbb{R}^+$, i.e., 
$Pr(y|x ; \theta, \sigma_1^2,..., \sigma_L^2) \propto \prod _{i=1}^L  \frac{1}{\sqrt{\sigma_i^2}}\exp(-\frac{\Vert y - \hat{y}_i\Vert_2^2 }{2 \sigma_i^2})$. 
Then we compute the empirical expected log-likelihood for a maximum likelihood estimator (MLE):
\begin{align}
 \hat{E}\big[\ln (Pr(y|x))\big]
\propto  \hat{E} \big[\sum _{i=1}^L( -\frac{\Vert y - \hat{y}_i\Vert^2_2}{\sigma_i^2} -  \ln \sigma_i^2 ) \big]
= \sum _{i=1}^L ( -\frac{\tilde{\ell}_i }{\sigma_i^2} -  \ln \sigma_i^2 ),
\label{eq:mle_to_log_barrier}
\end{align}
where $\hat{E}$ is averaging over samples, and $\tilde{\ell}_i$ is the empirical estimate of $\ell_i$. If we fix $\theta$ and optimize over $\sigma_i^2$, we get $\sigma_i^2 = \tilde{\ell}_i$.
As computing the empirical means is expensive over large data-sets, \adaloss replaces $\tilde{\ell}_i$ with $\hat{\ell}_i$, the exponential moving average of the losses, and sets $B_i \propto \hat{\ell}_i^{-1} \approx \sigma_i^{-2}$ so as to solve the MLE online by jointly updating $\theta$ and $B_i$. We note that the naturally appeared $\ln \sigma_i^2$ terms in Eq.~\ref{eq:mle_to_log_barrier} are log-barriers preventing $B_i=0$. Inspired by this observation, we form the following joint optimization over $\theta$ and $B_i$ for general losses without probability models:
\begin{align}
    \min _{\theta, B_1,...,B_L} \sum _{i=1}^L (B_i \ell _i(\theta) - \lambda \ln B_i),
    \label{eq:weighted_sum_log_barrier}
\end{align}
where $\lambda > 0$ is a hyper parameter to balance between the log-barriers and weighted losses. Under the optimal condition, $B_i=\frac{\lambda}{\ell_i}$. \adaloss estimates this with $B_i \propto \hat{\ell}_i(\theta)^{-1}$.  
We can also eliminate $B_i$ from Eq.~\ref{eq:weighted_sum_log_barrier} under the optimal condition, and we transform Eq.~\ref{eq:weighted_sum_log_barrier} to the following problem:
\begin{align}
    \min _{\theta} \sum _{i=1}^L \ln \ell _i (\theta).
    \label{eq:geometric_mean}
\end{align}
This is equivalent to minimizing the geometric mean of the expected training losses, and it differs from minimizing the expected geometric mean of losses, as $\ln$ and expectation are not commutable. 
Eq.~\ref{eq:geometric_mean} discards any constant scaling of losses automatically discarded as constant offsets, so that the scale difference between the early and late losses are automatically reconciled. Geometric mean is also known as the canonical mean to measure multiple positive quantities of various scales. To derive \adaloss directly from Eq.~\ref{eq:geometric_mean}, we note that the gradient of the objective in Eq.~\ref{eq:geometric_mean} is $\sum _{i=1}^L \frac{ \nabla \ell _i (\theta)} {\ell_i (\theta) }$, and gradient descent combined with \adaloss estimates the gradient with 
$\sum _{i=1}^L \frac{ \nabla \ell _i (\theta)} {\hat {\ell}_i (\theta) }$.


\section{Sequence of Exponentially Deepening Anytime Neural Networks (EANN)}
\label{sec:eann}

In practice, we often observe \anns using \adaloss to be much more competitive in their later half than the early half on validation sets, such as in Table.~\ref{tab:compare_f} of Sec.~\ref{sec:compare_opt}. Fortunately, we can leverage this effect to form competitive anytime predictors at every budget, with a constant fraction of additional computation. Specifically, we assemble \anns whose depths grow exponentially. Each \ann only starts computing if the smaller ones are finished, and its predictions are used if they are better than the best existing ones in validation. We call this ensemble an \textbf{EANN}, as illustrated in Fig.~\ref{fig:eann}. An EANN only delays the computation of any large \ann by at most a constant fraction of computation, because the earlier networks are exponentially smaller. Hence, if each \ann is near-optimal in later predictions, then we can achieve near-optimal accuracy at any test-time interruption, with the extra computation. 
Formally, the following proposition characterizes the exponential base and the increased computational cost.

\begin{proposition}
Let $b>1$. Assume for any $L$, any \ann of depth $L$ has competitive anytime prediction at  depth $i > \frac{L}{b}$ against the optimal of depth $i$. Then after $B$ layers of computation, EANN produces anytime predictions that are competitive against the optimal of depth $\frac{B}{C}$ for some $C > 1$, such that $\sup _B C = 2+ \frac{1}{b-1}$, and $C$ has expectation
$E_{B\sim uniform(1,L)}[C] \leq 1 - \frac{1}{2b} + \frac{1 + \ln (b)}{b-1}$.
\label{prop:eann}
\end{proposition}
This proposition says that an EANN is competitive at any budget $B$ against the optimal of the cost $\frac{B}{C}$. Furthermore, the stronger each anytime model is, i.e., the larger $b$ becomes, the smaller the computation inflation, $C$, is: as $b$ approaches $\infty$, $\sup  _B C$, shrinks to 2, and $E[C]$, shrinks to 1.
Moreover, if we have $M$ number of parallel workers instead of one, we can speed up EANNs by computing \anns in parallel in a first-in-first-out schedule, so that we effectively increase the constant $b$ to $b^M$ for computing $C$. It is also worth noting that if we form the sequence using regular networks instead of \anns, then we will lose the ability to output frequently, since at budget $B$, we only produce $\Theta(\log(B))$ intermediate predictions instead of the $\Theta(B)$ predictions in an EANN. We will further have a larger cost inflation, $C$, such that $\sup  _B C \geq 4$ and $E[C] \geq 1.5 + \sqrt{2} \approx 2.91$, so that the average cost inflation is at least about $2.91$.
We defer the proofs to the appendix.

\section{Experiments}
\label{sec:experiment_questions}

We list the key questions that our experiments aim to answer.
\begin{itemize}

\item How do anytime predictions trained with adaptive weights compare against those trained with static constant weights (over different architectures)?
(Sec.~\ref{sec:compare_opt})
\item How do underlying DNN architectures affect \anns? (Sec.~\ref{sec:compare_opt})
\item How can sub-par early predictions in \anns be mitigated by \ann ensembles? 
(Sec.~\ref{sec:eann_experiment})
\item How does data-set difficulty affect the adaptive weights scheme?
(Sec.~\ref{sec:weight_vs_dataset})
\end{itemize}

\subsection{Data-sets and Training Details}
\label{sec:exp}

\textbf{Data-sets.} We experiment on CIFAR10, CIFAR100~\citep{cifar}, SVHN~\citep{svhn}\footnote{Both CIFAR data-sets consist of 32x32 colored images. CIFAR10 and CIFAR100 have 10 and 100 classes, and each have 50000 training and 10000 testing images. We held out the last 5000 training samples in CIFAR10 and CIFAR100 for validation; the same parameters are then used in other models. We adopt the standard augmentation from \cite{supervisednet, resnet}.
SVHN contains around 600000 training and around 26032 testing 32x32 images of numeric digits from the Google Street Views. We adopt the same pad-and-crop augmentations of CIFAR for SVHN, and also add Gaussian blur.}
and ILSVRC~\citep{ILSVRC15}\footnote{
ILSVRC2012~\citep{ILSVRC15} is a visual recognition data-set containing around 1.2 million natural and 50000 validation images for 1000 classes. We report the top-1 error rates on the validation set using a single-crop of size 224x224, after scaling the smaller side of the image to 256, following~\citep{resnet}.}.

\textbf{Training details.} We optimize the models using stochastic gradient descent, with initial learning rate of 0.1, momentum of 0.9 and a weight decay of 1e-4. On CIFAR and SVHN, we divide the learning rate by 10 at 1/2 and 3/4 of the total epochs. We train for 300 epochs on CIFAR and 60 epochs on SVHN.  On ILSVRC, we train for 90 epochs, and divide the learning rate by 10 at epoch 30 and 60. We evaluate test error using single-crop.

\textbf{Base models.} We compare our proposed \adaloss weights against the intuitive \const weights. On CIFAR and SVHN, we also compare \adaloss against LINEAR and OPT, defined in Sec.~\ref{sec:multi_objective}. 
We evaluate the weights on multiple models including ResNet~\citep{resnet} and DenseNet~\citep{densenet}, and MSDNet~\citep{msdense}. For ResNet and DenseNet, we augment them with auxiliary predictors and losses, and call the resulting models Res\ann and Dense\ann, and defer the details of these models to the appendix Sec.~\ref{sec:implementation_ann}.

\subsection{Weight Scheme Comparisons}
\label{sec:compare_opt}

\begin{figure}
    \centering
    \resizebox{\textwidth}{!}{
    \subfloat[Relative Error Percentage Increases from the OPT]{
        \begin{tabular}{c|cccc}
        \hline
         & 1/4 & 1/2 & 3/4 & 1 \\
        \hline
        OPT
    	&  0.00 &  0.00 &  0.00 &  0.00 \\
        CONST
    	& \textbf{15.07} & 16.40 & 18.76 & 18.90 \\
        LINEAR
    	& 25.67 & 13.02 & 12.97 & 12.65 \\
        ADALOSS
     & 32.99 &  \textbf{9.97} &  \textbf{3.96} &  \textbf{2.73} \\
        \hline
        \end{tabular}
        \label{tab:compare_f}
    }
    ~
    \subfloat[Error Rates on ILSVRC]{
        \begin{tabular}{c|cccc}
        \hline
         & 1/4 & 1/2 & 3/4 & 1 \\
        \hline
        Res\annnp50+\const
    	& \textbf{54.34} & 35.61 & 27.23 & 25.14 \\
        Res\annnp50+\adaloss
    	& 54.98 & \textbf{34.92} & \textbf{26.59} & \textbf{24.42} \\
    	\hline
        Dense\annnp169+\const  
    	& 48.15 & 45.00 & 29.09 & 25.60 \\
        Dense\annnp169+\adaloss 
    	& \textbf{47.17} & \textbf{44.64} & \textbf{28.22} & \textbf{24.07} \\
        \hline      
        MSDNet38~\citep{msdense}
        & \textbf{33.9} & 28.0 & 25.7 & 24.3 \\
        MSDNet38+\adaloss 
        & 35.75 & 28.04 & 25.82 & \textbf{23.99} \\
        \hline
        \end{tabular}
        
        \label{tab:compare_f_ilsvrc}
    }
    }
    \caption{ \textbf{(a)} Average relative percentage increase in error from OPT on CIFAR and SVHN at 1/4, 1/2, 3/4 and 1 of the total cost. E.g., the bottom right entry means that if OPT has a 10\% final error rate, then \adaloss has about 10.27\%.
    \textbf{(b)} Test error rates at different fraction of the total costs on Res\annnp50 and Dense\annnp169. 
    }
    \vspace{-10pt}
\end{figure}

\textbf{\adaloss vs. \const on the same models.} Table~\ref{tab:compare_f} presents the average relative test error rate increase from OPT on 12  Res\anns on CIFAR10, CIFAR100 and SVHN\footnote{The 12 models are named by $(n,c)$ drawn from $\{ 7, 9, 13, 17, 25 \} \times \{ 16, 32 \}$ and $\{(9,64), (9,128)\}$, where $n$ represents the number of residual units in each of the three blocks of the network, and $c$ is the filter size of the first convolution.}. As training an OPT for each depth is too expensive, we instead report the average relative comparison at 1/4, 1/2, 3/4, and 1 of the total \ann costs. 
We observe that the \const scheme makes $15\sim 18\%$ more errors than the OPT, and the relative gap widens at later layers.  The \linear scheme also has about 13\% relative gap in later layers. In contrast, \adaloss enjoys small performance gaps in the later half of layers. 

\begin{figure}
    \centering
    
    \subfloat[Res\anns on CIFAR10]{
        \includegraphics[width=0.32\linewidth, keepaspectratio ]{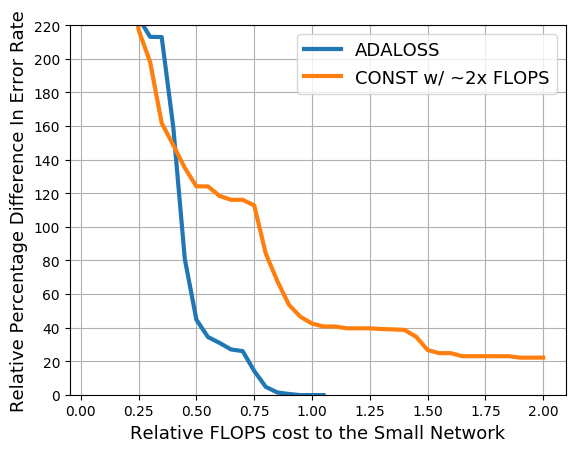}
        \label{fig:adaloss_vs_const_of_double_cost_cifar10}
    }
    ~
    \subfloat[Res\anns on CIFAR100]{
        \includegraphics[width=0.32\linewidth, keepaspectratio ]{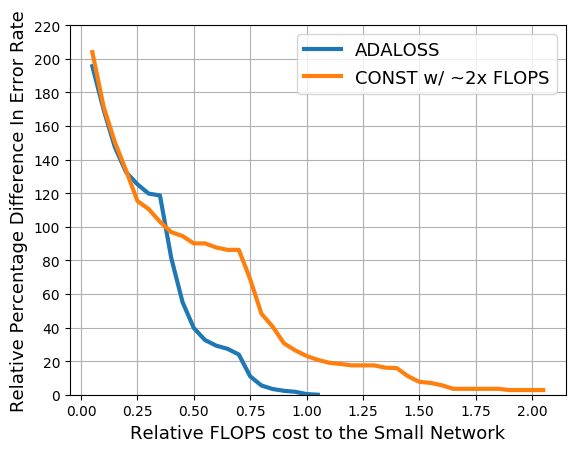}
        \label{fig:adaloss_vs_const_of_double_cost_cifar100}
    }
    ~
    \subfloat[Res\anns on SVHN]{
        \includegraphics[width=0.32\linewidth, keepaspectratio ]{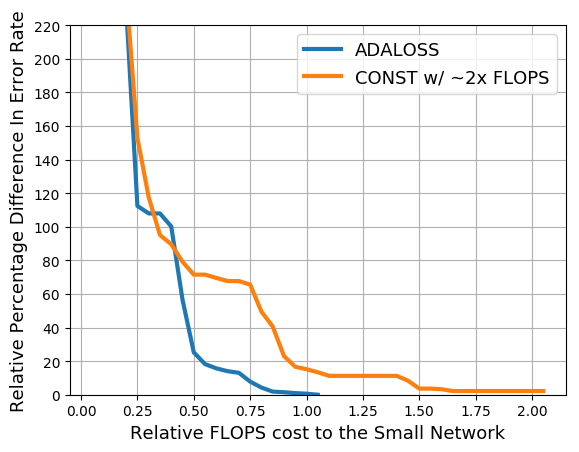}
        \label{fig:adaloss_vs_const_of_double_cost_svhn}
    }
    
    \subfloat[Res\anns on ILSVRC]{
        \includegraphics[width=0.32\linewidth, keepaspectratio]{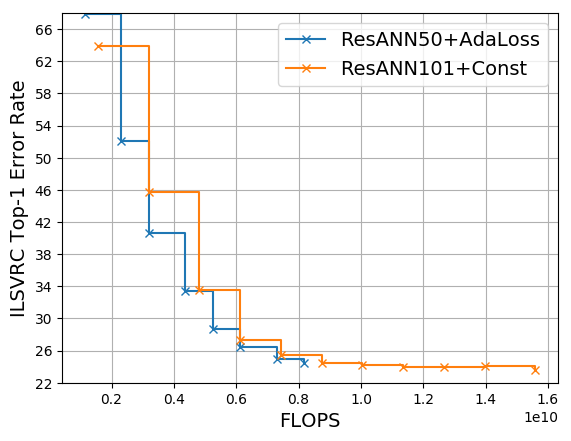}
        \label{fig:ilsvrc_adaloss_vs_const_of_double_cost}
    }
    ~
    \subfloat[MSDNet on ILSVRC]{
        \includegraphics[width=0.32\linewidth, keepaspectratio]{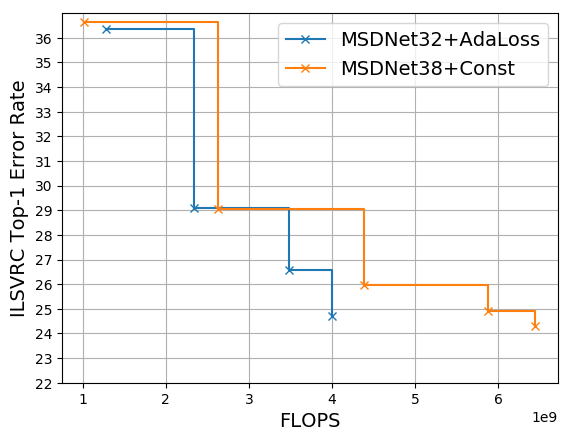}
        \label{fig:ilsvrc_adaloss_vs_const_of_double_cost_msdnet}    
    }
    ~
    \subfloat[\anns comparison on ILSVRC]{
        \includegraphics[width=0.32\linewidth, keepaspectratio]{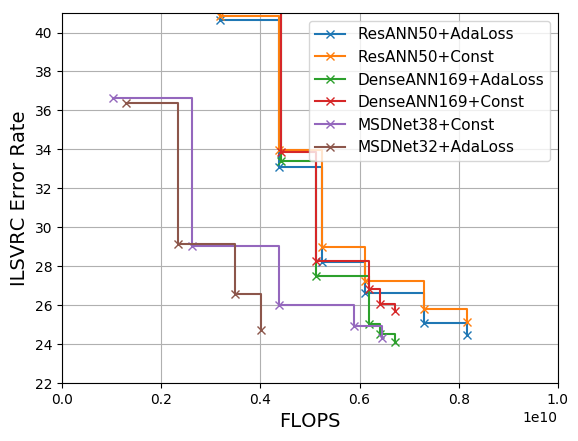}
        \label{fig:ilsvrc_compare_models}    
    }
    
    \caption{\textbf{(a-e)} Comparing small networks with \adaloss versus big ones using \const. With \adaloss, 
    the small networks achieve the same accuracy levels faster than large networks with \const. 
    \textbf{(f)} \anns performance are mostly decided by underlying models, but \adaloss is beneficial regardless models. }
    \label{fig:adaloss_vs_const_of_double_cost}
\end{figure}

On ILSVRC, we compare \adaloss against \const on Res\annnp50, Dense\annnp169, and MSDNet38, which have similar final errors and total computational costs (See Fig.~\ref{fig:ilsvrc_compare_models}). In Table~\ref{tab:compare_f_ilsvrc}, we observe the trade-offs between early and late accuracy on Res\annnp50 and MSDNet38. Furthermore, Dense\annnp169 performs \emph{uniformly} better with \adaloss than with \const. 

Since comparing the weight schemes requires evaluating \anns at multiple budget limits, and \adaloss and \const outperform each other at a significant fraction of depths on most of our experiments, we consider the two schemes \emph{incomparable on the same model}. However, our next experiments will show later predictions to be vastly more important than the early ones.

\textbf{Small networks with \adaloss vs. large ones with \const.} Practitioners may be interested in finding the smallest anytime models that can reach certain final accuracy thresholds, and unfortunately, the accuracy gain is often exponentially more costly as the accuracy saturates. To showcase the importance of this common phenomenon and its effect on choices of weight schemes, we compare \anns using \adaloss against \anns of about twice the cost but using \const. On CIFAR100, we average the relative comparison of six such pairs of Res\anns
\footnote{\adaloss takes $(n,c)$ from $\{7,9,13\} \times \{16, 32\}$, and \const takes $(n,c)$ from $\{13,17,25\} \times \{16, 32\}$.} in Fig.~\ref{fig:adaloss_vs_const_of_double_cost_cifar100}. E.g., the location (0.5, 200) in the plot means using half computation of the small \ann, and having 200\% extra errors than it. We observe small \anns with \adaloss to achieve the same accuracy levels faster than large ones with \const, because \const neglects the late predictions and large networks, and early predictions of large networks are not as accurate of those of a small ones. The same comparisons using Res\anns result in similar results on CIFAR10 and SVHN (Fig.~\ref{fig:adaloss_vs_const_of_double_cost_cifar10} and~\ref{fig:adaloss_vs_const_of_double_cost_svhn}). 
We also conduct similar comparisons on ILSVRC using Res\anns, and MSDNets, as shown in Fig.~\ref{fig:ilsvrc_adaloss_vs_const_of_double_cost} and Fig.~\ref{fig:ilsvrc_adaloss_vs_const_of_double_cost_msdnet}, and observe that the smaller networks with \adaloss can achieve accuracy levels faster than the large ones with \const, without sacrificing much final accuracy. 
For instance, MSDNet~\citep{msdense} is the state-of-the-art anytime predictor and is specially designed for anytime predictions, but by simply switching from their \const scheme to \adaloss, we significantly improve MSDNet32, which costs about 4.0e9 FLOPS (details in the appendix), to be about as accurate as the published result of MSDNet38, which has 6.6e9 total FLOPS in convolutions, and 72e6 parameters.

\textbf{Various base networks on ILSVRC.} We compare Res\anns, Dense\anns and MSDNets that have final error rate of near 24\% in Fig.~\ref{fig:ilsvrc_compare_models}, and observe that the anytime performance is mostly decided by the specific underlying model. Particularly, MSDNets are more cost-effective than Dense\anns, which in turn are better than Res\anns. 
However, \adaloss is helpful regardless of underlying model. Both Res\annnp50 and Dense\annnp169 see improvements switching from \const to \adaloss, which is also shown in Table~\ref{tab:compare_f_ilsvrc}. 
Thanks to \adaloss, Dense\annnp169 achieves the same final error using similar FLOPS as the original published results of MSDNet38~\citep{msdense}. This suggests that~\cite{msdense} improve over Dense\anns by having better early predictions without sacrificing the final cost efficiency via impressive architecture insight. Our \adaloss brings a complementary improvement to MSDNets, as it enables smaller MSDNets to reach the final error rates of bigger MSDNets, while having similar or better early predictions, as shown in the previous paragraph and Fig.~\ref{fig:ilsvrc_compare_models}. 




\subsection{EANN: Closing Early Performance Gaps by Delaying Final Predictions.}
\label{sec:eann_experiment}

\begin{figure}[t]
    \centering
    \subfloat[EANNs on CIFAR100]{
        \includegraphics[width=0.32\linewidth, keepaspectratio]{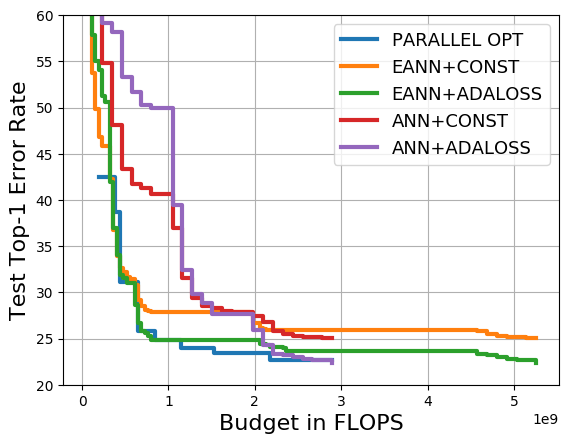}
        \label{fig:eann_f}
    }
    ~
    \subfloat[EANN on ILSVRC]{
        \includegraphics[width=0.32\linewidth, keepaspectratio]{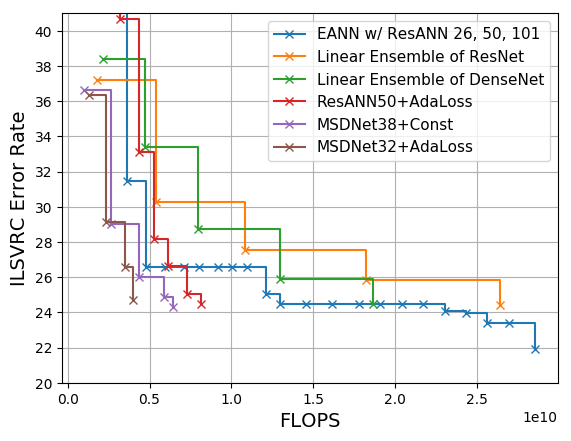}
        \label{fig:compare_ensemble}
    }
	~    
    \subfloat[Data-sets change \adaloss weights]{
        \includegraphics[width=0.32\linewidth, keepaspectratio]{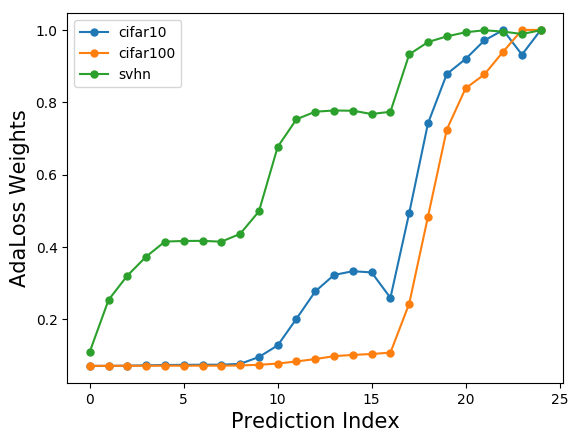}
        \label{fig:adaloss_weights}
    }
    \caption{ \textbf{(a)} EANN performs better if the \anns use \adaloss instead of \const. 
    \textbf{(b)} EANN outperforms linear ensembles of DNNs on ILSVRC.
	\textbf{(c)} The learned adaptive weights of the same model on three data-sets.
    }
\vspace{-15pt}
\end{figure}


\textbf{EANNs on CIFAR100.} In Fig.~\ref{fig:eann_f}, we assemble Res\anns to form EANNs\footnote{The Res\anns have $c=32$ and $n=7, 13, 25$, so that they form an EANN with an exponential base $b\approx 2$. 
By proposition~\ref{prop:eann}, the average cost inflation is $E[C]\approx 2.44$ for $b=2$, so that the EANN should 
compete against the OPT of $n=20$, using $2.44$ times of original costs.} on CIFAR100 and make three observations.
First, EANNs are better than the \ann in early computation, because the ensembles dedicate early predictions to small networks. Even though \const has the best early predictions as in Table~\ref{tab:compare_f}, it is still better to deploy small networks. 
Second, because the final prediction of each network is kept for a long period, \adaloss leads to significantly better EANNs than \const does, thanks to the superior final predictions from \adaloss. 
Finally, though EANNs delay computation of large networks, it actually appears closer to the OPT, because of accuracy saturation. Hence, EANNs should be considered when performance saturation is severe. 

\textbf{EANN on ILSVRC.}
\cite{msdense} and \cite{feedbacknet} use ensembles of networks of linearly growing sizes as baseline anytime predictors. However, in Fig.~\ref{fig:compare_ensemble}, an EANN using Res\anns of depths 26, 50 and 101 outperforms the linear ensembles of ResNets and DenseNets significantly on ILSVRC.
In particular, this drastically reduces the gap between ensembles and the state-of-the-art anytime predictor MSDNet~\citep{msdense}. 
Comparing Res\ann50 and the EANN, we note that the EANN achieves better early accuracy but delays final predictions. 
As the accuracy is not saturated by Res\ann26, the delay appears significant. Hence, EANNs may not be the best when the performance is not saturated or when the constant fraction of extra cost is critical.


\subsection{Data-set Difficulty versus Adaptive Weights}
\label{sec:weight_vs_dataset}

In Fig.~\ref{fig:adaloss_weights}, we plot the final \adaloss weights of the same Res\ann model (25,32) on CIFAR10, CIFAR100, and SVHN, in order to study the effects of the data-sets on the weights. We observe that from the easiest data-set, SVHN, to the hardest, CIFAR100, the weights are more concentrated on the final layers. This suggests that \adaloss can automatically decide that harder data-sets need more concentrated final weights to have near-optimal final performance, whereas on easy data-sets, more efforts are directed to early predictions. Hence, \adaloss weights may provide information for practitioners to design and choose models based on data-sets.

\vspace{-2pt}
\section{Conclusion and Discussion}
\vspace{-2pt}
This work devises simple adaptive weights, \adaloss, for training anytime predictions in DNNs. We provide multiple theoretical motivations for such weights, and show experimentally that adaptive weights enable small \anns to outperform large \anns with the commonly used non-adaptive constant weights. Future works on adaptive weights includes examining \adaloss for multi-task problems and investigating its ``first-order'' variants that normalize the losses by individual gradient norms to address unknown offsets of losses as well as the unknown scales. We also note that this work can be combined with orthogonal works in early-exit budgeted predictions~\citep{cascade_nn, adaptivenn} for saving average test computation. 


\section*{Acknowledgements}
This work was conducted in part through collaborative participation in the Robotics Consortium sponsored by the U.S Army Research Laboratory under the Collaborative Technology Alliance Program, Cooperative Agreement W911NF-10-2-0016. The views and conclusions contained in this document are those of the authors and should not be interpreted as representing the official policies, either expressed or implied, of the Army Research Laboratory of the U.S. Government. The U.S. Government is authorized to reproduce and distribute reprints for Government purposes notwithstanding any copyright notation herein.

\bibliographystyle{icml2018}
{\small
\bibliography{ann}
}

\clearpage
\appendix

\section{Sketch of Proof of Proposition~\ref{prop:eann}}
\begin{proof}
For each budget consumed $x$, we compute the cost $x'$ of the optimal that EANN is competitive against. The goal is then to analyze the ratio $C = \frac{x}{x'}$. 
The first ANN in EANN has depth 1. The optimal and the result of EANN are the same. Now assume EANN is on depth $z$ of ANN number $n+1$ for $n\geq 0$, which has depth $b^{n}$. \\
(Case 1) For $z \leq b^{n-1}$, EANN reuse the result from the end of ANN number $n$. 
The cost spent is $x = z + \sum _{i=0}^{n-1} b^i = z + \frac{b^n-1}{b-1}$. 
The optimal we compete has cost of the last ANN, which is $b^{n-1}$
The ratio satisfies:
\begin{align*} 
C &= x / x' = \frac{z}{b^{n-1}} + 1 + \frac{1}{b-1} - \frac{1}{b^{n-1}(b-1)} \\
&\leq 2 + \frac{1}{b-1} + \frac{1}{b^{n-1}(b-1)} \xrightarrow[]{n\rightarrow \infty} 2+ \frac{1}{b-1}. 
\end{align*}
Furthermore, since $C$ increases with $z$, 
\begin{align*}
&E_{z \sim Uniform(0, b^{n-1})}[C] \\
&\leq b^{1-n} \int _0 ^{b^{n-1}} 
    z b^{1-n}+ 1 + \frac{1}{b-1} \;dz \\
&= 1.5 + \frac{1}{b-1}.
\end{align*}
\\
(Case 2) For $b^{n-1} < z \leq b^n$, EANN outputs anytime results from ANN number $n+1$ at depth $z$. 
The cost is still $x = z +\frac{b^n-1}{b-1}$. The optimal competitor has cost $x' = z$.  Hence the ratio is 
\begin{align*}
C &= x/ x' = 1 + \frac{b^n-1}{z(b-1)} \\
&\leq 2 + \frac{1}{b-1} + \frac{1}{b^{n-1}(b-1)} \xrightarrow[]{n\rightarrow \infty} 2+ \frac{1}{b-1}.
\end{align*}
Furthermore, since $C$ decreases with $z$, 
\begin{align*}
&E_{z \sim Uniform(b^{n-1}, b^n)}[C] \\
& \leq (b-1)^{-1}b^{1-n} \Big[ ( 2 + \frac{1}{b-1}) \\
&\hspace{10pt} + \int _{b^{n-1}} ^{b^{n}} 
    2 + \frac{1}{b-1} + \frac{b^n -1}{z(b-1)} \; dz \Big] \\
& \xrightarrow[]{n\rightarrow \infty}
   1 + \frac{b\ln{b}}{(b-1)^2}
\end{align*}

Finally, since case 1 and case 2 happen with probability $\frac{1}{b}$ and $(1-\frac{1}{b})$, we have
\begin{align}
    \sup _B C &= 2+ \frac{1}{b-1} \\
\intertext{and}
    E_{B\sim Uniform(0, L)}[C] &\leq 1 - \frac{1}{2b} + \frac{1}{b-1} + \frac{\ln{b}}{b-1}.
\end{align}
We also note that with large $b$, $\sup _B C \rightarrow 2$ and $E[C] \rightarrow 1$ from above.
\end{proof}

If we form a sequence of regular networks that grow exponentially in depth instead of \ann, then the worst case happen right before a new prediction is produced. Hence the ratio between the consumed budget and the cost of the optimal that the current anytime prediction can compete, $C$, right before the number $n+1$ network is completed, is 
\[
    \frac{\sum _{i=1}^n b^i}{b^{n-1}} \xrightarrow[]{n\rightarrow \infty} \frac{b^2}{b-1} = 2 + (b-1) + \frac{1}{b-1} \geq 4. 
\]
Note that $(b-1) + \frac{1}{b-1} \geq 2$ and the inequality is tight at $b=2$. Hence we know $\sup _B {C}$ is at least 4. Furthermore, the expected value of $C$, assume $B$ is uniformly sampled such that the interruption happens on the $(n+1)^{th}$ network, is:
\begin{align*}
    E[C] &= \frac{1}{b^{n}} \int _0 ^{b^{n}} \frac{x + \frac{b^{n}-1}{b-1}}{b^{n-1}} \; dx \\ &\xrightarrow[]{n\rightarrow \infty} 1.5 + \frac{b-1}{2} + \frac{1}{b-1} \geq 1.5 + \sqrt{2} \approx 2.91.
\end{align*}
The inequality is tight at $b = 1 + \sqrt{2}$. With large $n$, since almost all budgets are consumed by the last few networks, we know the overall expectation $E_{B\sim Uniform(0, L)}[C]$ approaches $1.5 + \frac{b-1}{2} + \frac{1}{b-1}$, which is at least $1.5 + \sqrt{2}$.

\section{Additional Details of \adaloss for Experiments}

\label{sec:implementation_adaloss}

\textbf{Prevent tiny weights.} In practice, early $\hat{\ell}_i$ could be poor estimates of $\ell_i$, and we may have a feed-back loop where large losses incur small weights, and in turn, results in poorly optimized large losses.  To prevent such loops, we mix the adaptive weights with the constant weights. More precisely, we regularize Eq.~\ref{eq:geometric_mean} with the arithmetic mean of the losses:
\begin{align}
    \min _{\theta} \sum _{i=1}^L \big( \alpha (1 - \gamma) \ln \ell _i (\theta) + \gamma  \ell_i (\theta) \big),
    \label{eq:geometric_arithmetic_mean}
\end{align}
where $\alpha >0$ and $\gamma >0$ are hyper parameters. In practice, since DNNs often have elaborate learning rate schedules that assume $B_L=1$, we choose $\alpha  = \min_i \hat{\ell}_i(\theta)$ at each iteration to scale the max weight to 1. We choose $\gamma = 0.05$ from validation. Future works may consider more complex schemes where the weights start as constant weights and morph into \adaloss by gradually reducing $\gamma$ from 1 to 0.   

\textbf{Extra final weights.}  In our experiments, we often find that the penultimate layers have better accuracy relative to the OPT than the final layers on CIFAR, as suggested in Fig.~\ref{fig:loss_cifar100}. We believe this is because neighboring losses in an \ann are highly correlated, so that a layer can indirectly benefit from the high weights of its neighbors. The final loss is then at disadvantage due to its lack of successors. To remedy this, we can give the final loss extra weights, which turns the geometric mean in Eq.~\ref{eq:geometric_mean} into a weighted geometric mean. This is also equivalent to having a distribution of test-time interruption, where the interruption happens at all layers equally likely, except on the final layer. 
In our experiments, we do not use extra final weights on CIFAR10, CIFAR100 and SVHN to keep the weights simple, and we double the final weight on ILSVRC because the final accuracy there is critical for comparing against other non-anytime networks.

\section{Implementation Details of \anns}
\label{sec:implementation_ann}

\textbf{CIFAR and SVHN Res\anns.} For CIFAR10, CIFAR100~\citep{cifar}, and SVHN~\citep{svhn}, Res\ann follow \citep{resnet} to have three blocks, each of which has $n$ residual units. Each of such basic residual units consists of two 3x3 convolutions, which are interleaved by BN-ReLU. A pre-activation (BN-ReLU) is applied to the input of the residual units. The result of the second 3x3 conv and the initial input are added together as the output of the unit. The auxiliary predictors each applies a BN-ReLU and a global average pooling on its input feature map, and applies a linear prediction. The auxiliary loss is the cross-entropy loss, treating the linear prediction results as logits. For each $(n,c)$ pair such that $n < 25$, we set the anytime prediction period $s$ to be 1, i.e., every residual block leads to an auxiliary prediction. We set the prediction period $s=3$ for $n=25$.

\textbf{Res\anns on ILSVRC.} Residual blocks for ILSVRC are bottleneck blocks, which consists of a chain of 1x1 conv, 3x3 conv and 1x1 conv. These convolutions are interleaved by BN-ReLU, and pre-activation BN-ReLU is also applied. Again, the output of the unit is the sum of the input feature map and the result of the final conv. 
Res\annnp50 and 101 are augmented from ResNet50 and 101~\citep{resnet}, where we add BN-ReLU, global pooling and linear prediction to every two bottleneck residual units for ResNet50, and every three for ResNet101. 
We create Res\annnp26 for creating EANN on ILSVRC, and Res\annnp26 has four blocks, each of which has two bottleneck residual units. The prediction period is every two units, using the same linear predictors.

\textbf{Dense\anns on ILSVRC.} We augment DenseNet169~\citep{densenet} to create Dense\ann169. 
DenseNet169 has 82 dense layers, each of which has a 1x1 conv that project concatenation of previous features to $4k$ channels, where $k$ is the growth rate~\citep{densenet}, followed by a 3x3 conv to generate $k$ channels of features for the dense layer. The two convs are interleaved by BN-ReLU, and a pre-activation BN-ReLU is used for each layer. The 82 layers are organized into four blocks of size 6, 12, 32 and 32. Between each neighboring blocks, a 1x1 conv followed by BN-ReLU-2x2-average-pooling is applied to shrink the existing feature maps by half in the hight, width, and channel dimensions. We add linear anytime predictions every 14 dense layers, starting from layer 12 (1-based indexing). The original DenseNet paper~\citep{densenet} mentioned that they use drop-out with keep rate 0.9 after each conv in CIFAR and SVHN, but we found drop-out to be detrimental to performance on ILSVRC.

\textbf{MSDNet on ILSVRC.} MSDNet38 is described in the appendix of~\citep{msdense}. We set the four blocks to have 10, 9, 10 and 9 layers, and drop the feature maps of the finest resolution after each block as suggest in the original paper. 
We successfully reproduced the published results to 24.3\% error rate on ILSVRC using our Tensorflow implementation. We used the original published results for MSDNet38+\const in the main text. We use MSDNet32, which has four blocks of 6, 6, 10, and 10 layers, for the small network that uses \adaloss. We predict using MSDNet32 every seven layers, starting at the fourth layer (1-based indexing).

\end{document}